\documentclass[11pt,a4paper]{article}

\usepackage[hyperref]{emnlp2020}
\usepackage{times}
\usepackage{latexsym}

\usepackage{microtype}

\usepackage{url}
\usepackage{enumitem}
\usepackage{booktabs}
\usepackage{multirow}
\usepackage{graphicx}
\usepackage{array}
\usepackage{subcaption}
\usepackage{amsmath}
\usepackage{amsfonts}
\usepackage{amsthm,amssymb,amsopn,bm}
\usepackage{color,soul}
\usepackage{verbatim}
\usepackage{caption}
\usepackage{xcolor,colortbl}
\definecolor{green}{rgb}{0.1,0.1,0.1}
\usepackage{tabulary}
\usepackage{footnote}
\usepackage{blindtext}
\usepackage{algorithm}
\usepackage[noend]{algpseudocode}
\usepackage{stmaryrd}

\makesavenoteenv{tabular}
\makesavenoteenv{algorithm}
\usepackage{tabularx,booktabs}
\newcolumntype{Y}{>{\centering\arraybackslash}X}

\setlist{leftmargin=8mm}

\usepackage{tikz}
\def\checkmark{\tikz\fill[scale=0.4](0,.35) -- (.25,0) -- (1,.7) -- (.25,.15) -- cycle;}
\definecolor{gitred}{HTML}{FDB8C0}
\definecolor{gitgreen}{HTML}{006400}
\definecolor{chocolate}{HTML}{D2691E}
\definecolor{maroon}{HTML}{800000}
\definecolor{indigo}{HTML}{4B0082}
\definecolor{green}{HTML}{008000}
\definecolor{orange}{HTML}{fc8d62}
\definecolor{purple}{HTML}{8da0cb}

\newcolumntype{x}[1]{%
>{\raggedleft\hspace{0pt}}p{#1}}%

\aclfinalcopy 
\def\aclpaperid{1580} 

\providecommand{\sewon}[1]{
}

\newcommand\tf[1]{\textbf{#1}}

\newcolumntype{P}[1]{>{\centering\arraybackslash}p{#1}}

\newcommand\commentout[1]{}

\title{
    \taskname: Answering Ambiguous Open-domain Questions
}

\newcommand{\affilsup}[1]{\rlap{\textsuperscript{\normalfont#1}}}

\author{
    Sewon Min\affilsup{1,2},
    ~~~Julian Michael\affilsup{1},
    ~Hannaneh Hajishirzi\affilsup{1,3},
    ~~~Luke Zettlemoyer\affilsup{1,2}\\[0.25em]
    $^1$University of Washington \quad $^2$Facebook AI Research \quad $^3$Allen Institute for Artificial Intelligence \\
    \texttt{\{sewon,julianjm,hannaneh,lsz\}@cs.washington.edu} \\
}

\date{}

\begin{document}
\newcommand{\nq}{\textsc{Natural Questions}}
\newcommand{\nqopen}{\textsc{NQ-open}}
\newcommand{\triviaqa}{\textsc{TriviaQA}}
\newcommand{\trec}{\textsc{Trec}}
\newcommand{\webq}{\textsc{WebQuestions}}
\newcommand{\webqsplit}{\textsc{WebQuestions}}
\newcommand{\squad}{\textsc{SQuAD}}
\newcommand{\bert}{\textsc{BERT}}
\newcommand{\bart}{\textsc{BART}}
\newcommand{\tasknamelong}{Answering Ambiguous Open-domain Questions}
\newcommand{\taskname}{\textsc{AmbigQA}}
\newcommand{\dataname}{\textsc{AmbigNQ}}
\newcommand{\modelname}{\textsc{SpanSeqGen}}

\newcommand{\Prompt}{Prompt question}
\newcommand{\prompt}{prompt question}
\newcommand{\prompts}{prompt questions}
\newcommand{\noContext}{\textsc{Disambig-first}}
\newcommand{\dev}{development}
\newcommand{\devshort}{development}

\newcommand{\answerprediction}{multiple answer prediction}
\newcommand{\Answerprediction}{Multiple answer prediction}
\newcommand{\AnswerPrediction}{Multiple Answer Prediction}
\newcommand{\QD}{Question Disambiguation}
\newcommand{\qd}{question disambiguation}

\newcommand{\Dw}{$D_\textrm{partial}$}
\newcommand{\Dfhat}{$\hat{D}_\textrm{full}$}
\newcommand{\Df}{$D_\textrm{full}$}

\newcommand{\Fanswer}{$\mathrm{F1}_\text{ans}$}
\newcommand{\Fbleu}{$\mathrm{F1}_\text{BLEU}$}
\newcommand{\Fedit}{$\mathrm{F1}_\text{EDIT-F1}$}

\maketitle
\begin{abstract}
    Ambiguity is inherent to open-domain question answering; especially when exploring new topics, it can be difficult to ask questions that have a single, unambiguous answer.
    In this paper, we introduce \taskname, a new open-domain question answering task which involves
    finding every plausible answer, and then rewriting the question for each one to resolve the ambiguity.
    To study this task, we construct \dataname, a dataset covering 14,042 questions from \nqopen, an existing open-domain QA benchmark.
    We find that over half of the questions in \nqopen\ 
    are ambiguous, with diverse sources of ambiguity such as event and entity references.
    We also present strong baseline models for
    \taskname\
    which we show benefit from weakly supervised learning that incorporates \nqopen,
    strongly suggesting our new task and data will support significant future research effort.
    Our data and baselines are available at \url{https://nlp.cs.washington.edu/ambigqa}.
\end{abstract}

\section{Introduction}\label{sec:intro}
\begin{figure}[t]
\centering
\resizebox{\columnwidth}{!}{\includegraphics[width=\textwidth]{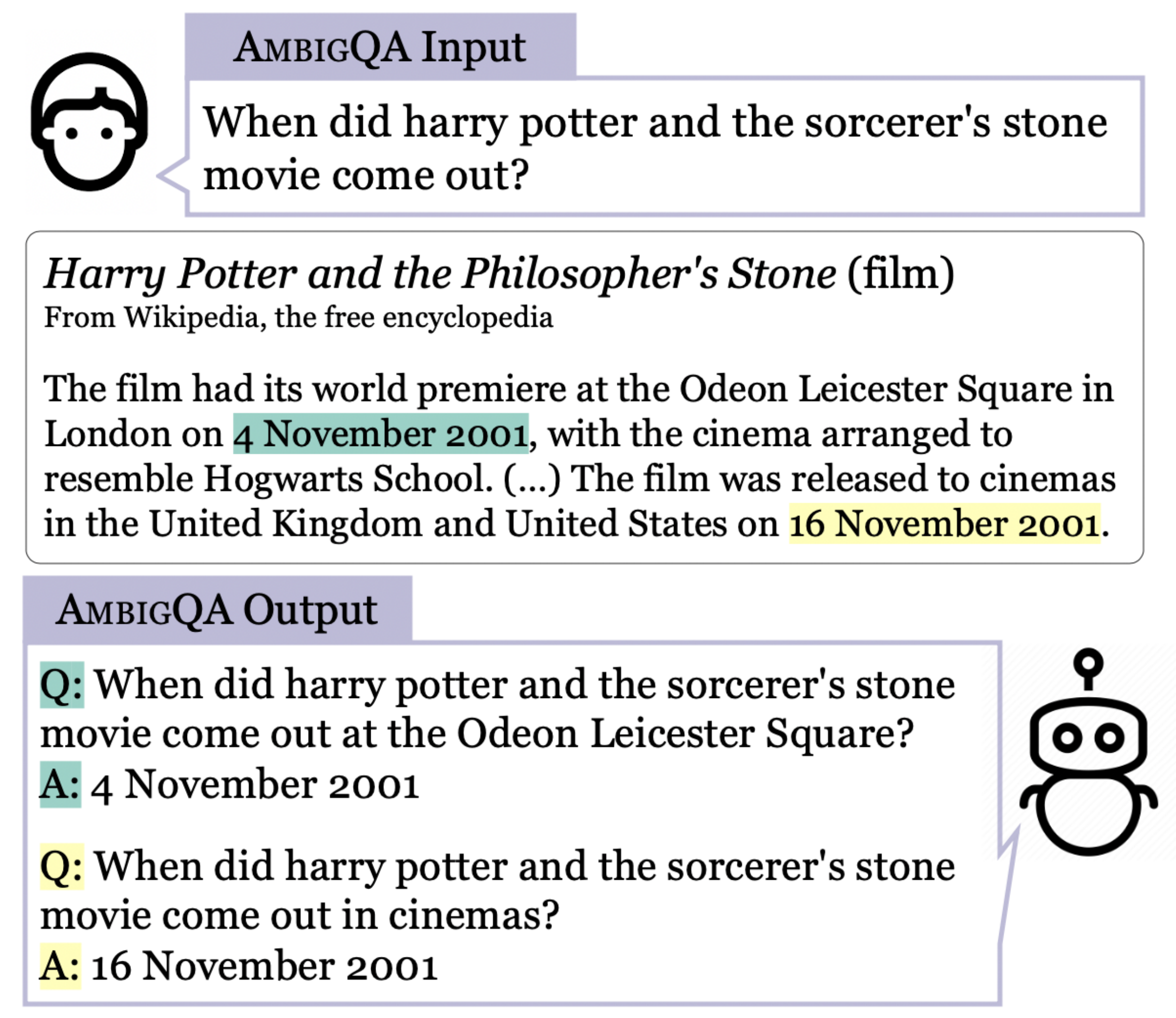}}
\caption{
    An \dataname\ example where
    the \prompt~(top) appears to have a single clear answer, but is actually ambiguous upon reading Wikipedia.
    \taskname\ requires producing the full set
    of acceptable answers while
    differentiating them from each other
    using disambiguated rewrites of the question.
}
\label{fig:intro}\end{figure}

In the open-domain setting, 
it can be difficult to formulate clear and unambiguous questions. 
For example, Figure~\ref{fig:intro} shows a Google search query~\cite{kwiatkowski2019natural} that, perhaps surprisingly, has two possible interpretations given the evidence in Wikipedia. Although open-domain question answering (QA) systems aim to answer any factoid question~\citep{voorhees1999trec}, existing methods assume questions have a single well-defined answer.
Nonetheless, ambiguity arises frequently in open-domain QA,
where questions are written during information gathering (e.g., search queries) without knowledge of the answer.
As we will see in Section~\ref{sec:data}, over 50\% of the questions we sampled from a set of Google search queries are ambiguous.
Furthermore, identifying ambiguities
is difficult both for humans and machines. As shown in Figure~\ref{fig:intro}, ambiguity is a function of both the question and the evidence provided by a large text corpus.

\begin{table*}[ht]
    \centering
    \footnotesize
    \begin{tabular}{l l}
    \toprule
        \tf{Type} & \tf{Example} \\
    \midrule
        \multirow{3}{*}{\shortstack[l]{Event references \\ (39\%)}} &
        What season does meredith and derek get married in grey's anatomy? \\
        & Q: In what season do Meredith and Derek get informally married in Grey's Anatomy? / A: Season 5 \\
        & Q: In what season do Meredith and Derek get legally married in Grey's Anatomy? / A: Season 7 \\
    \midrule
        \multirow{3}{*}{\shortstack[l]{Properties \\  (27\%)}} &
        How many episode in seven deadly sins season 2? \\
        & Q: How many episodes were there in seven deadly sins season 2, not including the OVA episode? / A: 25 \\
        & Q: How many episodes were there in seven deadly sins season 2, including the OVA episode? / A: 26 \\
    \midrule
        \multirow{3}{*}{\shortstack[l]{Entity references \\ (23\%)}} &
        How many sacks does clay matthews have in his career? \\
        & Q: How many sacks does Clay Matthews Jr. have in his career? / A: 69.5 \\
        & Q: How many sacks does Clay Matthews III have in his career? / A: 91.5 \\
    \midrule
        \multirow{3}{*}{\shortstack[l]{Answer types \\ (16\%)}} &
        Who sings the song what a beautiful name it is? \\
        & Q: Which group sings the song what a beautiful name it is? / A: Hillsong Live \\
        & Q: Who is the lead singer of the song what a beautiful name it is? / A: Brooke Ligertwood \\
    \midrule
        \multirow{4}{*}{\shortstack[l]{Time- \\ dependency \\ (13\%)}} &
        When does the new family guy season come out? \\
        & Q: When does family guy season 16 come out? / A: October 1, 2017 \\
        & Q: When does family guy season 15 come out? / A: September 25, 2016 \\
        & Q: When does family guy season 14 come out? / A: September 27, 2015 \\
    \midrule
        \multirow{3}{*}{\shortstack[l]{Multiple \\ sub-questions \\ (3\%)}} &
        Who was british pm and viceroy during quit india movement? \\
        & Q: Who was british viceroy during quit India movement? / A: Victor Hope \\
        & Q: Who was british pm during quit India movement? / A: Winston Churchill \\
    \bottomrule
    \end{tabular}
    \caption{
        Breakdown of the types of ambiguity in 100 randomly sampled items from the \dataname\ \dev\ data. Each example may fall into multiple categories.
    }
    \label{tab:ambiguity-breakdown}
\end{table*}

To study this challenge,
we introduce \taskname\
(\tasknamelong), 
a new task which involves disambiguating and answering potentially ambiguous questions. 
Specifically, the model must
(1) find a set of distinct, equally plausible answers to the question, and (2) provide minimal yet unambiguous rewrites of the question that clarify the interpretation which leads to each answer. Figure~\ref{fig:intro} shows two such disambiguated questions and their answers. 


To support the study of this task, we construct
a dataset called \dataname\ 
using 14,042 questions from an open-domain version of \nq~\citep{kwiatkowski2019natural}, denoted  \nqopen.
For each question,
annotators search for, navigate, and read
multiple Wikipedia pages to find
as many answers as possible.
The high prevalence of ambiguity makes the task difficult even for human experts; it is inherently difficult to know if you have found every possible interpretation of a question.
Nonetheless, we are able to collect high quality data 
covering high levels of ambiguity (2.1 distinct answers per question on average)
with high estimated agreement (89.0 F1)
on valid answers.
The types of ambiguity are diverse and sometimes subtle (Table~\ref{tab:ambiguity-breakdown}),
including ambiguous entity or event references,
or ambiguity over the answer type;
many are only apparent after
examining one or more Wikipedia pages.

To establish initial performance levels on this data, we present
a set of strong baseline methods. We extend a state-of-the-art QA model~\citep{karpukhin2020dense}
with three new components:
(1) set-based question answering with a sequence-to-sequence model,
(2) a question disambiguation model, and
(3) a modification to democratic co-training \citep{zhou2004democratic} which
leverages the partial supervision available in the full \nqopen\ dataset.
We also do an ablation study and qualitative analysis, which  suggest there is significant room for future work on this task. 
\commentout{We also do an extensive error analysis that shows there are significant gaps between baseline and human performance levels on both sub problems (including up to XX points even when using all the data), with question disambiguation being particularly challenging.}

To summarize, our contributions are threefold.
\begin{enumerate}[topsep=4pt,itemsep=6pt,partopsep=0pt, parsep=0pt]
    \item We introduce \taskname, a new task which requires identifying all plausible answers to an open-domain question, along with disambiguated questions to differentiate them.
    \item We construct \dataname, a dataset with 14,042 annotations on \nqopen\ questions containing diverse types of ambiguity.
    \item We introduce the first baseline models that produce multiple answers to open-domain questions, with experiments showing their effectiveness in learning from our data while highlighting avenues for future work.
\end{enumerate}

\section{Related Work}\label{sec:related}

\noindent
\textbf{Open-domain Question Answering} requires a system to answer any factoid question based on evidence provided by a large corpus such as Wikipedia~\citep{voorhees1999trec,chen2017reading}.
Existing benchmarks use questions of various types, from open-ended information-seeking~\citep{berant2013semantic,kwiatkowski2019natural,clark2019boolq} to more specialized trivia/quiz~\citep{joshi2017triviaqa,searchqa}. To the best of our knowledge, all existing formulations assume each question has a single clear answer.

Our work is built upon an open-domain version of \nq~\citep{kwiatkowski2019natural},
denoted \nqopen,
composed of questions posed by real users of Google search, each with an answer drawn from Wikipedia.
\nqopen\ has promoted several recent advances in open-domain question answering~\citep{lee2019latent,asai2019learning,min2019discrete,min2019knowledge,guu2020realm,karpukhin2020dense}.
Nonetheless, \citet{kwiatkowski2019natural} report that the answers to such questions are often debatable, and the average agreement rate on \nqopen\ test data is 49.2\%,\footnote{The \nqopen\ test data has 5-way annotations; we compute their pairwise agreement based on string match.} in large part due to ambiguous questions. In this work, we embrace this ambiguity as inherent to information seeking open-domain QA, and present the first methods for returning sets of answers paired with different interpretations of the question.

\vspace{0.5em}\noindent
\textbf{Clarification Questions}
have been used to study question ambiguity
in other settings.
Research on community Q\&A \citep{braslavski2017you,rao2018learning,rao2019answer}
studies finding underspecification in the question, but it does not find the answer to the original question.
In recent work,
\citet{xu2019asking} study clarification of questions that are intentionally annotated with pre-specified entity reference ambiguities.
\citet{aliannejadi2019asking} and \citet{zamani2020mimics} use clarification questions to refine intents of simple query logs without
immediately apparent information needs (e.g., single keywords like \textit{dinosaur}\footnote{The average query length in \citet{zamani2020mimics} is 2.6.}).
\sewon{Added Zamani et al.}

In contrast, we study open-domain factoid questions asked by real users: these present clear information needs, but carry diverse naturally occurring ambiguities (see Table~\ref{tab:ambiguity-breakdown}).
Furthermore, instead of prolonging the user's information-seeking session with clarification questions, our task formulation provides a complete and immediate solution with unambiguous rewrites of the original question.

\vspace{0.5em}\noindent
\textbf{Question Rewriting} is a novel, well-defined task which we propose for differentiating distinct answers.
To the best of our knowledge, it has not been studied for resolving ambiguity;
we are only aware of \citet{elgohary2019can} which use question rewriting to convert conversational questions into self-contained questions.

\section{Task: \taskname}\label{sec:task}\subsection{\taskname\ Setup}\label{subsec:setup}


Figure~\ref{fig:intro} depicts the \taskname\ task.
The input is a \prompt\ $q$, and the output is a list of $n$ question-answer pairs $(x_1, y_1), \dots, (x_n, y_n)$, where each $y_i$ is an equally plausible answer to $q$, and each $x_i$ is a minimally edited modification of $q$ whose answer is unambiguously $y_i$.  We consider two subtasks.

\paragraph{\AnswerPrediction.} Given a question $q$,  output a set of semantically distinct and equally plausible answers $y_1, \dots, y_n$, where
$n$ is unknown.

\paragraph{\QD.} Given $q$ and a set of
answers $y_1, \dots, y_n$, generate \textit{disambiguated questions}
$x_1, \dots, x_n$, where
each $x_i$ is a \textit{minimal edit}
of $q$ which makes it unambiguous
so that $y_i$ is a correct answer and
all $y_j$ for all $j \neq i$ are incorrect. 
When $n = 1$, this task is trivial,
as $x_1 = q$.



We choose to represent ambiguity with a set of disambiguated questions because it is well-defined, immediately human-interpretable, and allows for straightforward annotation of a wide range of ambiguities without complex guidelines.



\subsection{Evaluation Metrics}\label{subsec:metrics}

To evaluate model performance, we present several ways to compare a model prediction with $m$ question-answer pairs $(x_1, y_1), \dots, (x_m, y_m)$
with a gold reference set with $n$ pairs
$(\bar{x}_1, \bar{\mathcal{Y}}_1), \dots, (\bar{x}_n, \bar{\mathcal{Y}}_n)$.
Since there may be more than one way
to refer to a single answer
(e.g., {\em Michael Jordan} and {\em Michael Jeffrey Jordan})
each gold answer $\bar{\mathcal{Y}}_i$
is a set of acceptable answer strings,
where all $\bar{\mathcal{Y}}_i$ are disjoint.

We assign each predicted question-answer pair $(x_i, y_i)$ a \textit{correctness score} based on a string similarity function $f$ valued in $[0, 1]$.
$$c_i=\max\limits_{1 \leq j \leq n} \mathbb{I}[y_i \in \bar{\mathcal{Y}}_j] f(x_i, \bar{x}_j).$$
Intuitively, $c_i$ considers (1) the correctness of the answer and (2) the similarity
$f(x_i, \bar{x}_j)$
between the predicted and reference question.
We calculate F1 treating the $c_i$ as measures of correctness: \begin{gather*}
        \mathrm{prec}_{f} =
          \frac{\sum_i c_i}{m}, \qquad
        \mathrm{rec}_{f} =
          \frac{\sum_i c_i}{n}, \\
        \mathrm{F1}_{f} = \frac{2 \times  \mathrm{prec}_{f} \times \mathrm{rec}_{f}}{\mathrm{prec}_{f} + \mathrm{rec}_{f}}.
    \end{gather*}

\noindent

We consider three choices of $\mathrm{F}_{f}$.
\textbf{\Fanswer}
is the F1 score on answers only,
where $f$ always yields 1.
This may be used without the
question disambiguation step.
\textbf{\Fbleu}
accounts for string similarity between questions,
calculating $f$ with \textsc{BLEU}~\citep{papineni2002bleu}.
\textbf{\Fedit}
uses \textsc{Edit-F1} as $f$, where
\textsc{Edit-F1}
is a new measure that
represents each
disambiguated question 
by its added and deleted unigrams
compared to the \prompt,
and computes the F1 score between them.
For example, consider
the \prompt\ ``Who made the play the crucible?",
the reference ``Who wrote the play the crucible?"
and the prediction ``Who made the play the crucible in 2012?". 
The gold edits\footnote{Represented
as multisets, written using
$\lbag \texttt{bag} \rbag$
notation.} here are
$\lbag \colorbox{red!20}{\texttt{-made}},
       \colorbox{green!20}{\texttt{+wrote}} \rbag$
while the predicted edits are
$\lbag \colorbox{green!20}{\texttt{+in}},
       \colorbox{green!20}{\texttt{+2012}} \rbag$.
Their \textsc{Edit-F1} is thus zero, even though the questions are similar.
Unlike BLEU which we use to directly measure similarity to the gold question, this metric only gives credit for getting the key semantic differences correct between the original question and the clarification.

\section{Data: \dataname}\label{sec:data}\subsection{Data Collection}\label{subsec:data-collection}

We construct \dataname\ using \prompts\ from \nqopen\ and English Wikipedia as the evidence corpus.
We use Amazon Mechanical Turk
for crowdsourcing.

The crucial annotation challenge is maximizing recall: finding all possible distinct answers to a question.
This is difficult, as ambiguities are often only apparent after carefully searching the evidence for multiple possible answers.
However, we can collect high quality data with high levels of ambiguity using
careful worker selection and a two stage pipeline: \textit{generation} and \textit{validation}.


\paragraph{Generation.}
Workers in the first stage are given a \prompt\ and a search box that uses the Google Search API
restricted to English Wikipedia.
Allowing annotators to find Wikipedia pages on their own closely approximates the real process people use to answer open-ended questions---an approach with no existing large-scale dataset.\footnote{
    For instance, answers in \nqopen\ are annotated over pre-specified Wikipedia pages from the Google search engine.
}


Workers find all plausible answers to the question; when there are multiple, each answer is paired with a minimal edit of the \prompt\ which differentiates it from the other answers, in line with our task requirements.
A distinct answer may be annotated as multiple possible spans (e.g., {\em Michael Jordan} and {\em Michael Jeffrey Jordan}).
As a special case, some questions contain \textit{temporal deixis} which depends on the time of writing, e.g., ``When does the new family guy season come out?".
To avoid unmanageably many answers, we instruct workers to remove the time-dependence by rewriting the \prompt\ for
up to three most recent events before Jan 1, 2018,
e.g.,  ``When does family guy season 16 come out?" (see Table~\ref{tab:ambiguity-breakdown}).

\begin{table}[t]
    \centering
    \footnotesize
    \begin{tabular}{lrrrrr}
        \toprule
            \multirow{2}{*}{Split} & \multirow{2}{*}{\# data} &  \multicolumn{4}{c}{\# QAs \%} \\
            \cmidrule(lr){3-6}
             & & 1 & 2 & 3 & 4+ \\
        \midrule
            Train & 10,036 & 53 & 24 & 14 & 10 \\
            Dev & 2,002 & 49 & 23 & 14 & 13 \\
            Test & 2,004 & 44 & 24 & 16 & 16 \\
        \bottomrule
    \end{tabular}
    \caption{
	    Data statistics. For the number of QA pairs (\# QAs), the minimum is taken when there are more than 1 accepted annotations.
	}\label{tab:data-statistics}
\end{table}
\begin{figure*}
	\begin{subfigure}{.6\columnwidth}
        \centering
        \includegraphics[width=\columnwidth,trim=1mm 1mm 1mm 1mm,clip]{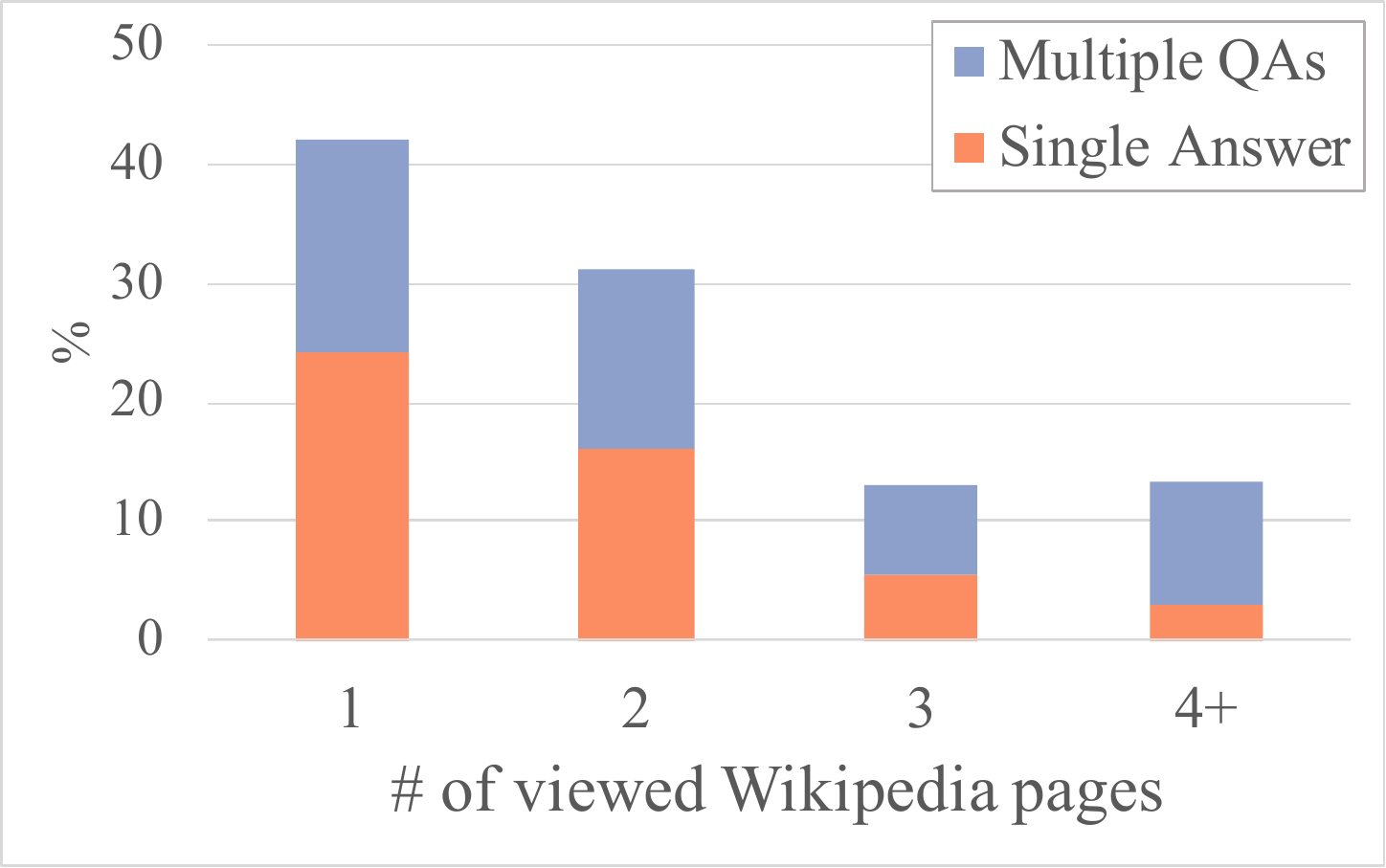}
        \caption{
            Number of unique Wikipedia pages visited by crowdworkers.$^\dagger$
        }
        \label{fig:data-analysis-pages}
    \end{subfigure}\hfill
    \begin{subfigure}{.6\columnwidth}
        \centering
        \includegraphics[width=\columnwidth,trim=1mm 1mm 1mm 1mm,clip]{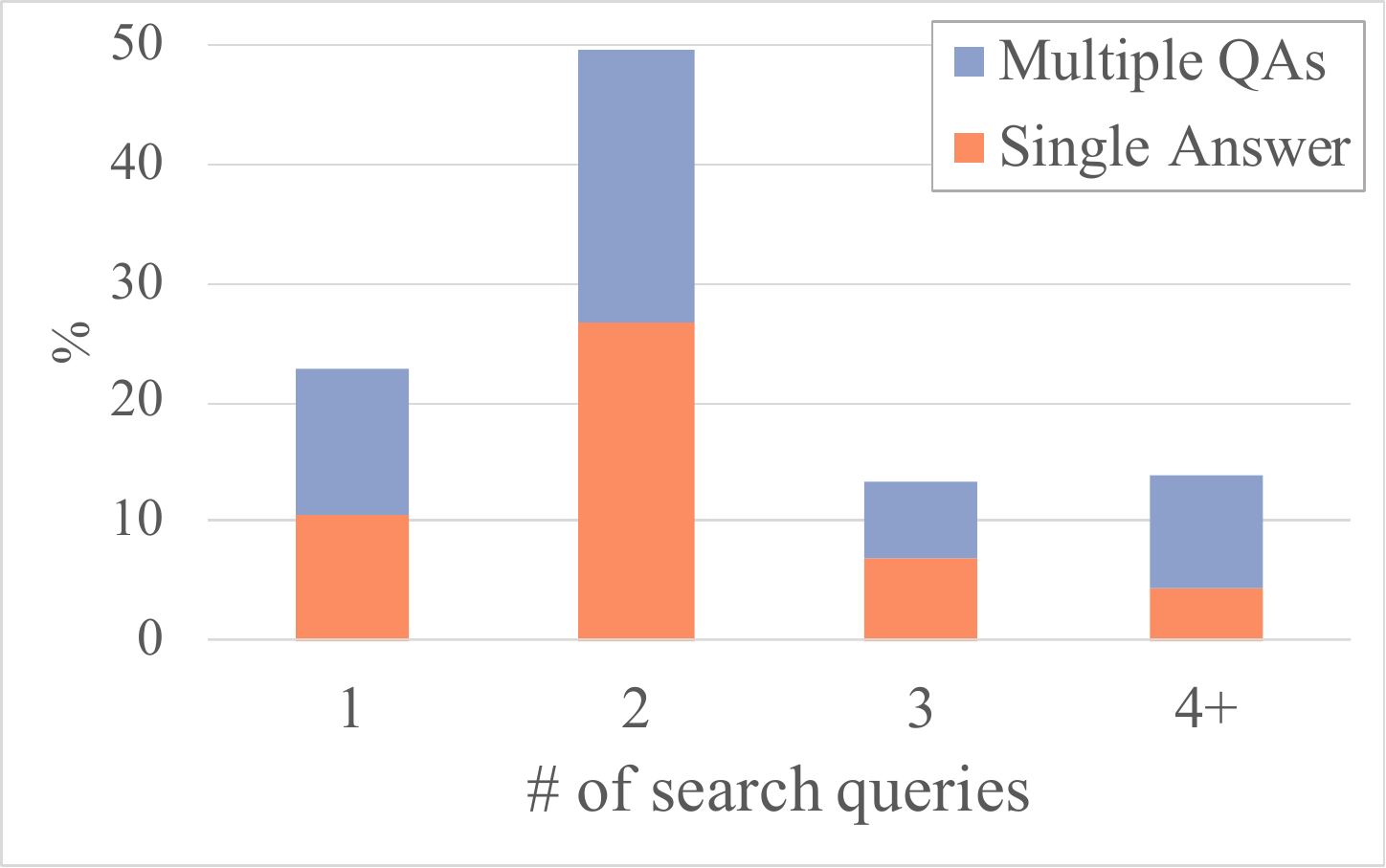}
        \caption{
            Number of search queries written by crowdworkers.
        }
        \label{fig:data-analysis-queries}
    \end{subfigure}\hfill
    \begin{subfigure}{.7\columnwidth}
        \centering
        \includegraphics[width=\columnwidth,trim=0 5cm 0 5cm,clip]{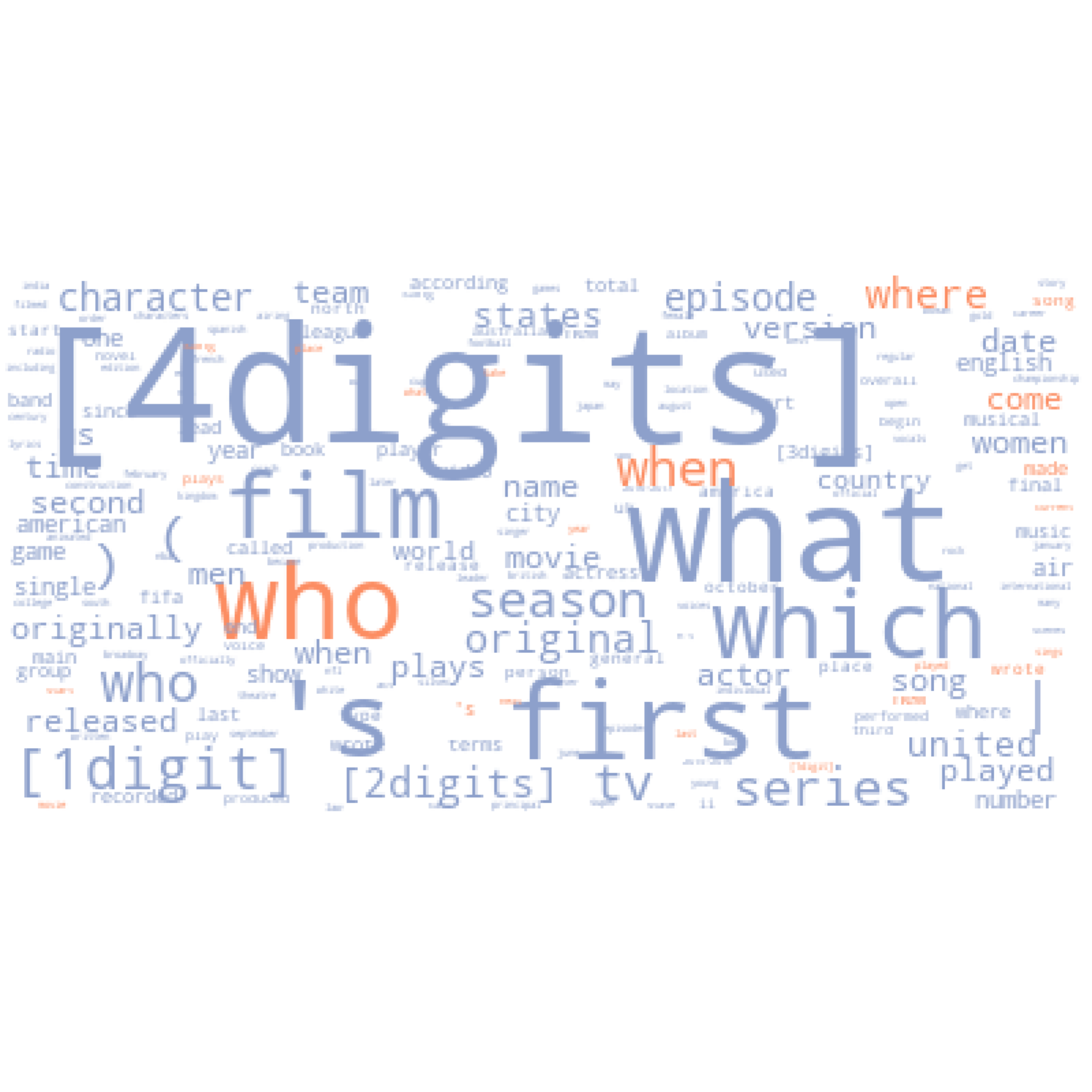}
        \caption{
            Word cloud of the edits made in questions; {\protect\color{purple}{$\blacksquare$}} and {\protect\color{orange}{$\blacksquare$}} indicate added and deleted unigrams, respectively. 
        }
        \label{fig:data-analysis-wordcloud}
    \end{subfigure}
	\caption{Data Analysis on the \dev\ data.
	{$^\dagger$This is actually an underestimate; we could not track when annotators viewed pages by following hyperlinks for technical reasons.}
	} \label{fig:data-analysis}
\end{figure*}

\paragraph{Validation.} Workers in the validation stage review the annotations provided by multiple generators.
Validators mark each generator's annotations as correct or incorrect, or provide a new set of question-answer pairs by combining the valid ones from each generator.
They search Wikipedia as generators do, and are additionally given Wikipedia pages that generators viewed to speed up the process.
Validation is skipped when annotated answers from all generators exactly match (37\% of cases).


\paragraph{Quality control.}
We recruit highly qualified workers through a qualification test (details in Appendix~\ref{app:data-collection}).
Although the task was difficult for most workers,
we found that our highly qualified full-time workers, given quick and detailed feedback on their work, produced high accuracy and recall.
For \dev\ and test data, we use two generators and one validator per \prompt.
For training data, we skip validation and only use one generator per question.

\paragraph{Inter-annotator agreement.}
\commentout{
    In 37\% of cases, all generators produced the exact same set of answers in string match.
    In another 39\%, all annotations from the generation stage were accepted in the validation stage.
    In the remaining 24\% of cases, the validator chose one generator's annotations or wrote a custom combined set of question-answer pairs, because some generators either missed some plausible answers or included invalid answers.}
Evaluating generators against each other on the \dev\ set yields 60.8 \Fanswer.
All annotations passed validation for 76\% of questions, while validators made changes (edits or exclusions) in the remaining 24\%.
The average \Fanswer\ between co-authors and workers on a sample of 50 validations was 89.0\%.
This indicates that, despite the intrinsic difficulty and subjectivity of the task, humans agree on the boundary between valid and invalid answers in most cases.

\subsection{Data Analysis}\label{subsec:data-analysis}

The final dataset contains 14,042 annotated examples, split consistently with \nqopen.
As shown in Table~\ref{tab:data-statistics}, over 50\% of \dev\ and test examples contain multiple question-answer pairs. This indicates a high rate of ambiguity in \nqopen, even though previous work has studied it with the assumption that each question has a single answer.
We also find a discrepancy between \dev\ and test; this is likely due to the way in which \nqopen\ is constructed, which over-samples difficult questions in the test set (see Appendix~\ref{app:dev-and-test} for details).
The training set contains relatively fewer ambiguous examples (47\%), presumably because using only one worker per training example yielded slightly lower recall.

\paragraph{Types of ambiguity.}
Table~\ref{tab:ambiguity-breakdown} shows a breakdown of the types of ambiguity in \dataname.
They are diverse, including ambiguity in entity references, event references, properties, and answer types, with a relatively uniform distribution between them. 
In comparison to \citet{xu2019asking}, who intentionally elicit questions with ambiguous entity references, our analysis shows that \textit{unintended} ambiguity comes from diverse sources.
In many cases, ambiguity is not apparent from the \prompt\ alone, but only after researching the question on Wikipedia, as evidenced by differences in model performance (Section~\ref{subsec:main-results}).

\paragraph{Annotator behavior.}
Figures~\ref{fig:data-analysis-pages} and~\ref{fig:data-analysis-queries} show the number of unique Wikipedia pages
and the number of search queries used by workers during annotation.
More often than not, workers used multiple queries and navigated multiple Wikipedia pages,
showing how our setup captures ambiguity in the \textit{retrieval} step of open-domain question answering, which is missed in approaches that assume a pre-specified evidence document.

\paragraph{Distribution of edits.}
Figure~\ref{fig:data-analysis-wordcloud} shows unigram edits made to questions in the \dev\ data, where we remove stopwords except wh-words and group numeric values by the number of digits.
Adding numerals such as years is common, as they can easily disambiguate entity or event references or remove time dependence.
Wh-word changes are also common, especially for specifying the answer type (e.g., from {\em who} to {\em which group}; see Table~\ref{tab:ambiguity-breakdown}).
The distribution of edits is fairly long-tailed, with the 100 most frequent edits covering 36\% of the total, and the top 1,000 covering 69\%.

\section{Model}\label{sec:model}
To set initial performance levels on \dataname, we present a baseline \taskname\ model combining ideas from recent advances in open-domain QA~\cite{karpukhin2020dense} and generation~\cite{lewis2019bart}. 
Given a \prompt\ $q$, our model predicts answers $y_1..y_n$, and generates corresponding questions $x_1..x_n$ conditioning on $q$, the answers $y_1..y_n$, and the evidence passages.
A novel co-training step also allows the model
to leverage the partial supervision available in \nqopen.

\paragraph{\AnswerPrediction.}
Here we describe \modelname, our model for \answerprediction.
Following \citet{karpukhin2020dense}, a state-of-the-art model on \nqopen, \modelname\ first retrieves 100 passages with a \bert-based~\citep{bert} dual encoder, and reranks them using a BERT-based cross encoder.
Then, instead of predicting an answer span from the top 1 passage as \citet{karpukhin2020dense} does, \modelname\ uses another sequence-to-sequence model based on \bart~\citep{lewis2019bart}. Specifically, it conditions on the concatenation of $q$ and the top passages in order up to 1024 tokens, and sequentially generates distinct answers token-by-token, separated by [\texttt{SEP}].
We pretrain \modelname\ on \nqopen\ and finetune it on \dataname.

\begin{algorithm}
\caption{
    Democratic co-training with weak supervision (Section~\ref{subsec:model-cotraining}).
}\label{alg:cotraining}
\begin{algorithmic}[1]
\footnotesize
\State {\protect\color{gitgreen}  {\textit{// Each question in \Df\ has an answer list annotated}}} 
\State {\protect\color{gitgreen}  {\textit{// Each question in \Dw\ has one answer annotated}}} 
\State \Dfhat\ $\gets$ \Df  
\For{$iter \in \{1..N \}$}
    \State {\protect\color{gitgreen}  {\textit{// Train $C$ sequence-to-sequence QA models}}}
    \For{$i \in \{1..C \}$}
        \State $\phi_i \gets train($\Dfhat) 
    \EndFor
    \State $\hat{D}_L \gets $\Df  
    \For{$(q^j, y^j) \in $\Dw}
        \State {{\protect\color{gitgreen}\textit{// Get predictions by using $y_j$ as prefix}}}
        \State $\hat{Y^j} \gets \{\hat{y} \mid \hat{y}\neq y^j, $ and
        \State  $\qquad |\{i \mid \hat{y} \in \phi_i(q^j|y^j), 1\leq i \leq C\}| > \frac{C}{2}$ 
        \State $\}$ 
        \If{$|\hat{Y}^j|>0$}
            \State {\protect\color{gitgreen}      {\textit{// Add it as a multiple answer case}}}
            \State \Dfhat$ \gets \hat{D}_L \cup \{(q^j, \{y^j\} \cup \hat{Y^j})\}$ 
        \ElsIf{$\forall i=1..C, |\phi_i(x^j)-\{y^j\}|=0$}
            \State {\protect\color{gitgreen}  {\textit{// Add it as a single answer case}}} 
            \State \Dfhat$ \gets \hat{D}_L \cup \{(q^j, \{y^j\})\}$
        \EndIf
    \EndFor
\EndFor
\end{algorithmic}
\end{algorithm}

We develop \modelname\ primarily because \citet{karpukhin2020dense} is designed for generating a single answer, but \modelname\ also boosts the performance on \nqopen\ (41.5$\rightarrow$42.2 on the test data).
We include ablations on different approaches and models in Section~\ref{subsec:main-results}.

\paragraph{\QD.}
We design a \qd\ (QD) model based on \bart. The model generates each question $x_i$ ($i=1..n$) conditioning on the concatenation of $q$, the target answer $y_i$, other answers $y_1..y_{i-1}, y_{i+1}..y_{n}$, and the top passages as used by \modelname.
We pretrain on \nqopen\ to generate questions given an answer and passage, and then finetune it on the full task data in \dataname.
We include ablations on different variants of the model in Section~\ref{subsec:main-results}.

\begin{table*}
	\centering \footnotesize
	\begin{tabular}{l P{0.7cm} P{0.7cm} P{0.7cm} P{0.7cm} P{0.7cm} P{0.7cm} P{0.7cm} P{0.7cm}}
    \toprule
        \multirow{2}{*}{Model} & \multicolumn{2}{c}{\Fanswer\ (\em all)} &  \multicolumn{2}{c}{\Fanswer\ (\em multi)} & \multicolumn{2}{c}{\Fbleu} & \multicolumn{2}{c}{\Fedit} \\
        \cmidrule(lr){2-3} \cmidrule(lr){4-5} \cmidrule(lr){6-7} \cmidrule(lr){8-9} 
        & {dev} & {test} & {dev} & {test} &{dev} & {test} & {dev} & {test} \\
    \midrule
        \noContext & 28.1 & 24.8 & 21.9 & 18.8 & 4.2 & 4.0 & 2.7 & 2.2 \\
        Thresholding + QD & 37.1 & 32.3 & 28.4 & 24.8 & 13.4 & 11.3 & 6.6 & 5.5 \\
        \modelname\ + QD & 39.7 & 33.5 & 29.3 & 24.5 & 13.4 & 11.4 & 7.2 & 5.8 \\
    \midrule
        \modelname$^\dagger$ + QD & 41.2 & 35.2 & 29.8 & 24.5 & 13.6 & 10.6 & 7.4 & 5.7 \\ 
        \modelname$^\dagger$ (Co-training) + QD & \tf{42.3} & \tf{35.9} & \tf{31.7} & \tf{26.0} & \tf{14.3} & \tf{11.5} & \tf{8.0} & \tf{6.3} \\ 
    \bottomrule
    \end{tabular}
	\caption{
	    Results on \dataname.
	    The {\em multi} measure only considers examples with multiple question-answer pairs.
	    $^\dagger$ indicates ensemble.
	    See Appendix~\ref{app:dev-and-test} for details on the discrepancy between \dev\ and test.
	}\label{tab:result}
\end{table*}

\paragraph{Co-training with weak supervision.}\label{subsec:model-cotraining}
Given the prevalence of unlabelled ambiguity in \nqopen, we introduce a method that treats the \nqopen\ annotations as weak supervision and learns to discover potential ambiguity in the data.
We modify a democratic co-training algorithm~\citep{zhou2004democratic} as described in Algorithm~\ref{alg:cotraining}. 
We iteratively grow the training set \Dfhat\ from \dataname\ (\Df) with silver data from \nqopen\ (\Dw) predicted by a majority of a set $C$ of \modelname\ models trained on \Dfhat.
The key step is injecting the known answer $y^j$ from \nqopen\ as a prefix to \modelname's output during prediction.
In each step, if a majority of $C$ predict an additional answer, we assume we have found a false negative and add the result to the training set \Dfhat.
If all models predict no additional answer, we add the example to \Dfhat\ with $y^j$ as a single answer. 

\section{Experiments}\label{sec:exp}
\begin{table*}[th]
	\centering \footnotesize
	\begin{tabular}{lccc P{1.2cm} P{1.2cm} P{1.2cm} P{1.2cm} }
    \toprule
        \multirow{2}{*}{Model} &
        \multirow{2}{*}{$q$} &
        \multirow{2}{*}{$y_i$} &
        \multirow{2}{*}{\shortstack[l]{$y_1..y_{i-1},$ \\ $y_{i+1}..y_{n}$}} & \multicolumn{2}{c}{{\em Full task}} &  \multicolumn{2}{c}{{\em Gold answers given}} \\
        \cmidrule(lr){5-6} \cmidrule(lr){7-8}
        &&& & \Fbleu & \Fedit & \Fbleu & \Fedit \\
    \midrule
        QD model & \checkmark & \checkmark & \checkmark & 14.3 & \tf{8.0} & 40.1 & \tf{19.2} \\
        - \prompt & - & \checkmark & \checkmark & 6.7 & 7.7 & 15.1 & \tf{19.2}  \\
        - untargeted answers & \checkmark & \checkmark & - & 14.2 & 7.3 & 41.2 & 17.2 \\
        Always \prompt & \checkmark & - & - & \tf{15.9} & 0.0 & \tf{47.4} & 0.0 \\
    \bottomrule
    \end{tabular}
	\caption{
	    Ablations on \qd\ (\dev\ data, multiple answers only).
	    QD model refers to the \qd\ model described in Section~\ref{sec:model}.
	    For \answerprediction, we use \modelname$^\dagger$ with co-training ({\em Full task}) or the gold answers ({\em Gold answers given}).
	}\label{tab:qg-ablation}
\end{table*}

\begin{table*}
    \centering
    \footnotesize
    \setlength\tabcolsep{3.8pt}
    \begin{tabular}{l}
    \toprule
        \textbf{\Prompt\ \#1:} Where was snow white and the huntsman filmed? \\
        \textbf{Reference:} \\
        Q: Where were beach scenes for snow white and huntsman predominantly filmed? / A: Marloes Sands Beach \\
        Q: Where was principal photography for snow white and huntsman filmed? / A: United Kingdom \\
        Q: Where was castle in snow white and huntsman filmed? / A: Gateholm island \\
        \textbf{Prediction of \noContext:} (\Fanswer=0.40, \Fedit=0.00) \\
        Q: Where was snow white and the huntsman filmed in 2017? / A: Marloes Sands Beach \\
        Q: Where was snow white and the huntsman filmed during the filming of Season 1 of the TV series? / A: Marloes Sands Beach \\
        \textbf{Prediction of \modelname:} (\Fanswer=0.80, \Fedit=0.69) \\
        Q: Where was snow white and huntsman principal photography filmed / A: United Kingdom \\
        Q: Where were beach scenes for snow white and huntsman mostly filmed / A: Marloes Sands Beach \\
    \midrule
        \textbf{\Prompt\ \#2:} When was the city of new york founded? \\
        \textbf{Reference:} \\
        Q: When was city of new york founded by dutch and initially called new amsterdam? / A: 1624 \\
        Q: When was city of new york under english control and renamed to new york? / A: 1664 \\
        \textbf{Prediction of \modelname:} (\Fanswer=1.00, \Fedit=0.67) \\
        Q: When was city of new york city founded with dutch protection? / A: 1624 \\
        Q: When was city of new york city founded and renamed with english name? / A: 1664 \\
    \bottomrule
    \end{tabular}
    \caption{
        Model predictions on samples from the \dev\ data.
        \tf{(\#1)} 
        \noContext\ generates questions that look reasonable on the surface but don't match the facts.
        \modelname\ produces the reasonable answers and questions, although not perfect.
        \tf{(\#2)} \modelname\ produces correct answers and questions.
    }
    \label{tab:model-analysis}
\end{table*}

\begin{table}
    \centering
    \footnotesize
    \setlength\tabcolsep{3.8pt}
    \begin{tabular}{lc}
    \toprule
        \textbf{Reference has multiple answers} \\
        Multiple answer prediction is correct & 2\% \\
        Multiple answer prediction is partially correct$^\dagger$ & 40\% \\
        Multiple answer prediction is  incorrect & 14\% \\
    \midrule
        \textbf{Reference has one answer}\\
        Over-generated predictions & 2\% \\
        Correct single answer prediction & 26\% \\
        Incorrect single answer prediction & 12\% \\
    \midrule
        \textbf{Reference is incorrect} & 4\% \\
    \bottomrule
    \end{tabular}
    \caption{
        Analysis of predictions made by \modelname\ with co-training, on 50 samples from the \dev\ data.
        Examples shown in Appendix (Table~\ref{tab:appendix-pred-analysis}).\\
        {\footnotesize $^\dagger$In 15 out of 20 cases, the model generates only one answer.}
    }
    \label{tab:pred-analysis}
\end{table}


We describe the baseline models used in our experiments, followed by results and ablations.
Implementation details and hyperparameters of all models are provided in Appendix~\ref{app:impl-details}.

\subsection{Baselines}\label{subsec:baselines}

\paragraph{\noContext.} This baseline disambiguates the \prompt\  without any context from plausible answers or reference passages. Specifically, it implements the following pipeline:
(1) Feed the \prompt\ $q$ into a \bert-based binary classifier to determine whether it is ambiguous.
(2) If $q$ is ambiguous, pass it into a \bart-based model which generates a sequence of disambiguated questions $x_1..x_n$ ($n>1$), separated by \texttt{[SEP]}; otherwise, consider only $x_1 = q$.
(3) Feed each $x_i$ into a state-of-the-art model on \nqopen~\citep{karpukhin2020dense} to produce its answer $y_i$. 

\paragraph{Thresholding + QD.}
We also include a model based on \citet{karpukhin2020dense}, with thresholding for \answerprediction\ and our \qd\ (QD) model.
\citet{karpukhin2020dense} outputs a likelihood score for each span; we obtain ${y}_1..{y}_n$ by taking valid spans with likelihood larger than a hyperparameter $\gamma$.
The model is trained to
maximize the marginal likelihood of
any span in the gold answer set
$\bar{\mathcal{Y}}_1..\bar{\mathcal{Y}}_n$.
As with \modelname, we pretrain on \nqopen\ and finetune on \dataname.
We then produce disambiguated questions using our \bart-based QD model (Section~\ref{sec:model}).

\subsection{Results}\label{subsec:main-results}

Table~\ref{tab:result} reports the performance of our baselines; example model outputs are provided in Table~\ref{tab:model-analysis}.

\paragraph{Main results.}
We first find that \noContext\ is significantly worse than other models.
In particular, classification accuracy on whether the \prompt\ is ambiguous is 67\%, close to the majority baseline (60\%).
When the model does identify an ambiguous question, its rewrites often look reasonable on the surface, but do not match the facts.
For instance, in example 1 of Table~\ref{tab:model-analysis}, it asks about filming in 2017 and during season 1 for \textit{Snow White and the Huntsman}, which was actually a film released in 2012.
This shows that reading evidence documents is crucial for identifying and characterizing ambiguities.



While \modelname\ outperforms \citet{karpukhin2020dense} with thresholding, the difference is not as great as we expected.
This suggests two things.
First, thresholding may be a surprisingly effective baseline for outputting multiple answers, even though the answers must compete with each other for probability mass in order to surpass the threshold $\gamma$.
Second, maximizing likelihood in a sequence-to-sequence model like \modelname\
may not produce well-calibrated results.
For instance, the model seems to suffer due to variation in the length of the output sequence, outputting shorter sequences on average (3.0 tokens) than gold (6.7).\footnote{This problem has also been reported in other conditional generation tasks~\citep{sountsov-sarawagi-2016-length,stahlberg2019nmt}; we leave it for future work.}
This leads to low recall when there are multiple answers; our best model achieves a precision of $49.6$ and recall of $25.3$ for its \Fanswer\ of $31.7$ on such questions.

Overall, \modelname\ achieves reasonable \Fanswer\ scores. \Fanswer\ on examples with multiple question-answer pairs ({\em multi}) are lower, indicating that predicting all plausible answers is more challenging than predicting a single answer, as expected.
\modelname\ also obtains the best performance in \Fbleu\ and \Fedit, although their absolute values are low in general; we discuss this in our question disambiguation ablations below.

There is a substantial difference in performance between \dev\ and test overall, likely due to distributional differences in the original questions in \nqopen; detailed discussion is in Appendix~\ref{app:dev-and-test}.

\paragraph{Effect of co-training.}
The last two rows of Table~\ref{tab:result} reports the effect of our co-training method.
As co-training requires multiple trained models, we compare with a naive ensemble.
While we see gains from ensembling alone, an ensemble trained with the co-training method achieves the best performance on all metrics. This result demonstrates the potential of jointly using \dataname\ and partial supervision from \nqopen.

\paragraph{Ablations on \qd.}
Table~\ref{tab:qg-ablation} reports results of an ablation experiment on \qd\ (QD). Among our ablations, we include models without the \prompt\ or untargeted answers as input, and a naive baseline that always outputs the \prompt.
We report the metrics both in the scenarios of the full task and the gold answers given, to see the performance dependent on and independent from multiple answer prediction, respectively.\footnote{
    Note that a high \Fanswer\ and low \Fedit\ may not indicate bad \qd. For instance, if a model correctly predicts one out of two answers and does not perform any edits to the question, it obtains high \Fanswer\ and zero \Fedit, despite the error being in answer prediction.
}

Simply copying the \prompt\ gives high \Fbleu, which is natural since the questions were disambiguated using minimal edits. 
This justifies using \Fedit\ to evaluate semantic differences from the \prompt.
In addition, we find that our QD model conditioned on all available context is better than other variants in overall metrics. 

Performance is low overall, even given the gold answers, highlighting the challenge of the task.
We think there are two major reasons.
First, maximizing the likelihood of the output sequence can miss the importance of \textit{edits} to the \prompt, leading the QD model to miss the information that is most important to differentiate one answer from the others.
Second, there is a lack of annotated data, especially for \qd\ which does not benefit from weakly supervised learning with \nqopen; 
future work can explore how to maximize the use of supervision from other available data.
It is also worth noting that the metric may miss edits that are semantically correct, but phrased differently (see Table~\ref{tab:model-analysis}, example 2).

\begin{table}
    \centering \footnotesize
	\begin{tabular}{l P{1.05cm} P{0.6cm} P{0.6cm} }
    \toprule
        \multirow{2}{*}{Model} &
        {\scriptsize \nqopen} EM & \Fanswer\ (\em all) & \Fanswer\ (\em multi) \\
    \midrule
        \multicolumn{4}{l}{\textbf{\em Dev}} \\
        \citet{min2019knowledge} & 34.7 & 30.8 & 20.4 \\ 
        \citet{asai2019learning} & 31.7 & 29.7 & 19.7 \\ 
        \citet{karpukhin2020dense} & 39.8 & 35.2 & \tf{26.5} \\ 
        \modelname & \tf{42.0} & \tf{36.4} & 24.8 \\ 
    \midrule
        \multicolumn{4}{l}{\textbf{\em Test}} \\
        \citet{min2019knowledge} & 34.5 & 27.5 & 17.0 \\ 
        \citet{asai2019learning} & 32.6 & 27.9 & 17.7 \\ 
        \citet{karpukhin2020dense} & {41.5} & {30.1} & \tf{23.2} \\ 
        \modelname & \tf{42.2} & \tf{30.8} & 20.7 \\ 
    \bottomrule
    \end{tabular}
	\caption{
	    Zero-shot performance on \answerprediction\ of the models trained on \nqopen.
	    We report Exact Match (EM) on \nqopen\ and \Fanswer\ on \dataname.
	}\label{tab:zero-shot-result}
\end{table}

\subsection{Zero-shot results}
Since \dataname\ provides an evaluation set with explicit sets of multiple answers, we can also test if models trained on partial supervision only (\nqopen) are capable of producing full answer sets.
\sewon{Added.}
In fact, the problem of ambiguity already exists in previous QA tasks, and a single labeled answer can be viewed as a sample from a multi-modal distribution of answers.
This setting is important for modeling in domains where single-answer datasets are available but full annotations like in \dataname\ are not.
To this end, we present a zero-shot setting where a system predicts multiple distinct answers without using \dataname\ training data.
We include four \nqopen\ models including ours, consisting of diverse approaches and model architectures, as baselines.
These models, when trained on \nqopen, may be made to predict multiple answers via thresholding as described in Section~\ref{subsec:baselines}.\footnote{We allow using \dev\ data to tune the threshold $\gamma$, although this arguably makes our setting not zero-shot in the strictest sense.} 
Table~\ref{tab:zero-shot-result} reports zero-shot performance.
Although \modelname\ outperforms \citet{karpukhin2020dense} in the standard setting, it is worse in zero-shot \Fanswer\ ({\em multi}),
potentially because thresholding exacerbates the problems that \modelname\ has with long sequences (Section~\ref{subsec:main-results}).

\subsection{Error Analysis}\label{subsec:analysis}

Table~\ref{tab:pred-analysis} reports an analysis of predictions by \modelname\ with co-training, based on 50 random samples from the \dev\ data; examples can be found in the Appendix (Table~\ref{tab:appendix-pred-analysis}).
When there are multiple reference answers, the model rarely gets all correct answers, although often generates a subset of them.
In 15 out of 20 partially correct cases,
the model produces only one answer,
consistent with the under-generation we found in Section~\ref{subsec:main-results}.
In four out of those 15 cases, the model prediction is arguably the most likely answer,\footnote{For example, a question ``Who did \texttt{<title-of-the-} \texttt{song>}'' is ambiguous, but a well-known performer of the song may be argued to be a more likely answer than its little-known songwriter.}
but in the other 11 cases, it hard to argue for one answer over the other(s).
It is also worth noting that accuracy on examples with a single answer is quite high, being correct in 13 out of 20 cases. 
This estimated accuracy on unambiguous questions is higher than state-of-the-art levels on \nqopen\ ($42$ EM), suggesting that \nqopen\ may substantially underestimate performance due to the prevalence of unmarked ambiguity.
Together with our experimental results, this seems to indicate that recall of multiple answers is one of the primary challenges in AmbigQA.

\section{Conclusion \& Future Work
}\label{sec:concl}We introduced \taskname, a new task that involves providing multiple possible answers to a potentially ambiguous open-domain question, and providing a disambiguated question corresponding to each answer.
We constructed \dataname, a dataset with 14,042 annotations on \nqopen\ questions.
Our analysis shows the dataset contains diverse types of ambiguity, often not visible from the \prompt\ alone.
We also introduced a first baseline model for producing multiple answers to open-domain questions, with experiments showing its effectiveness in learning from our data while highlighting possible areas for improvement. 
\sewon{Edited this sentence and added the following paragraph.}

Future research developing on AmbigQA models may include
explicitly modeling ambiguity over events and entities or in the retrieval step, as well as improving performance on the difficult problems of answer recall and question disambiguation.
Furthermore, future work may build on the AmbigQA task with more open-ended approaches such as (1) applying the approach to QA over structured data (such as ambiguous questions that require returning tables), (2) handling questions with no answer or ill-formed questions that require inferring and satisfying more complex ambiguous information needs, and (3) more carefully evaluating usefulness to end users.

\section*{Acknowledgments}
This research was supported by ONR N00014-18-1-2826, DARPA N66001-19-2-403,  the NSF (IIS-1252835, IIS-1562364), an Allen Distinguished Investigator Award, and the Sloan Fellowship.
We thank Mandar Joshi, H2Lab members and the anonymous reviewers for their helpful comments and suggestions.

\bibliography{journal-abbr,bib}
\bibliographystyle{acl_natbib}

\clearpage
\appendix
\section{Data Collection Details}\label{app:data-collection}We use Amazon Mechanical Turk\footnote{\url{www.mturk.com}} and Spacro~\citep{michael2018crowdsourcing}\footnote{\url{github.com/julianmichael/spacro}} for crowdsourcing.
All data was collected in February and March of 2020.
We use the Google Search API\footnote{\url{developers.google.com/custom-search/}} restricted to English Wikipedia for the search tool.

\begin{figure*}
\centering
    \begin{subfigure}{2\columnwidth}\resizebox{\columnwidth}{!}{\includegraphics[width=\textwidth]{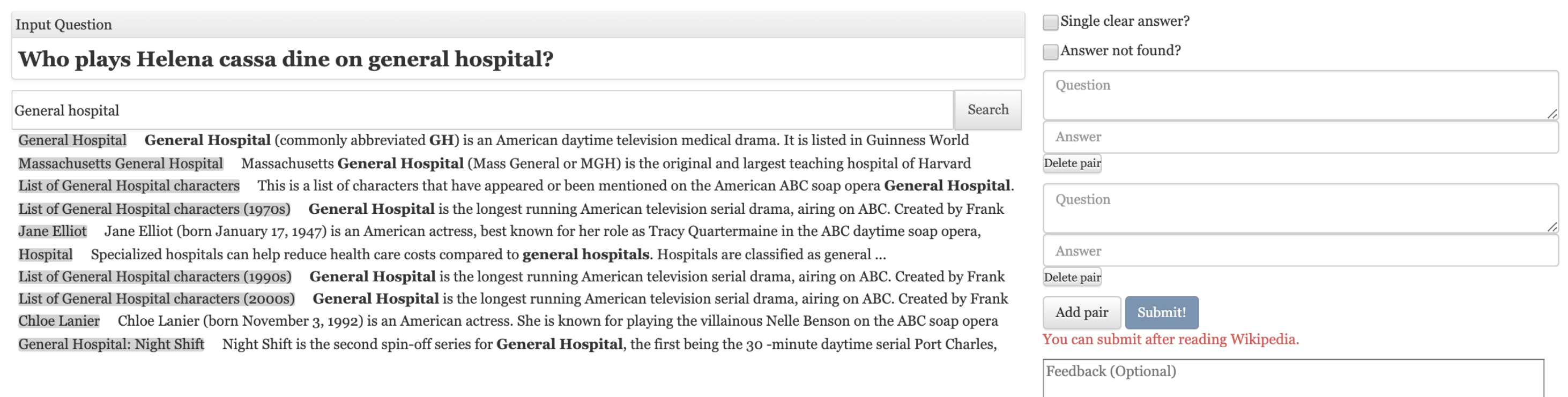}}\caption{
        Interface in the generation stage when the workers write a query and see the search results.
    }\label{fig:interface-1}\end{subfigure}
    \begin{subfigure}{2\columnwidth}\resizebox{\columnwidth}{!}{\includegraphics[width=\textwidth]{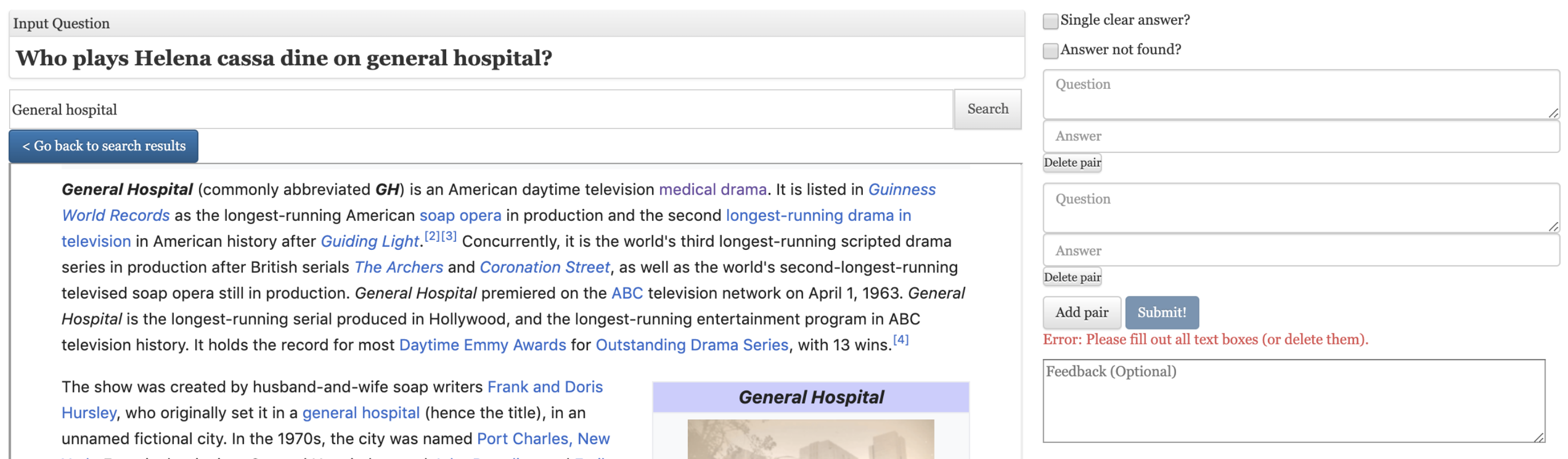}}\caption{
        Interface in the generation stage when the workers click and read one of Wikipedia pages from the search results.
    }\label{fig:interface-2}\end{subfigure}
    \begin{subfigure}{2\columnwidth}\resizebox{\columnwidth}{!}{\includegraphics[width=\textwidth]{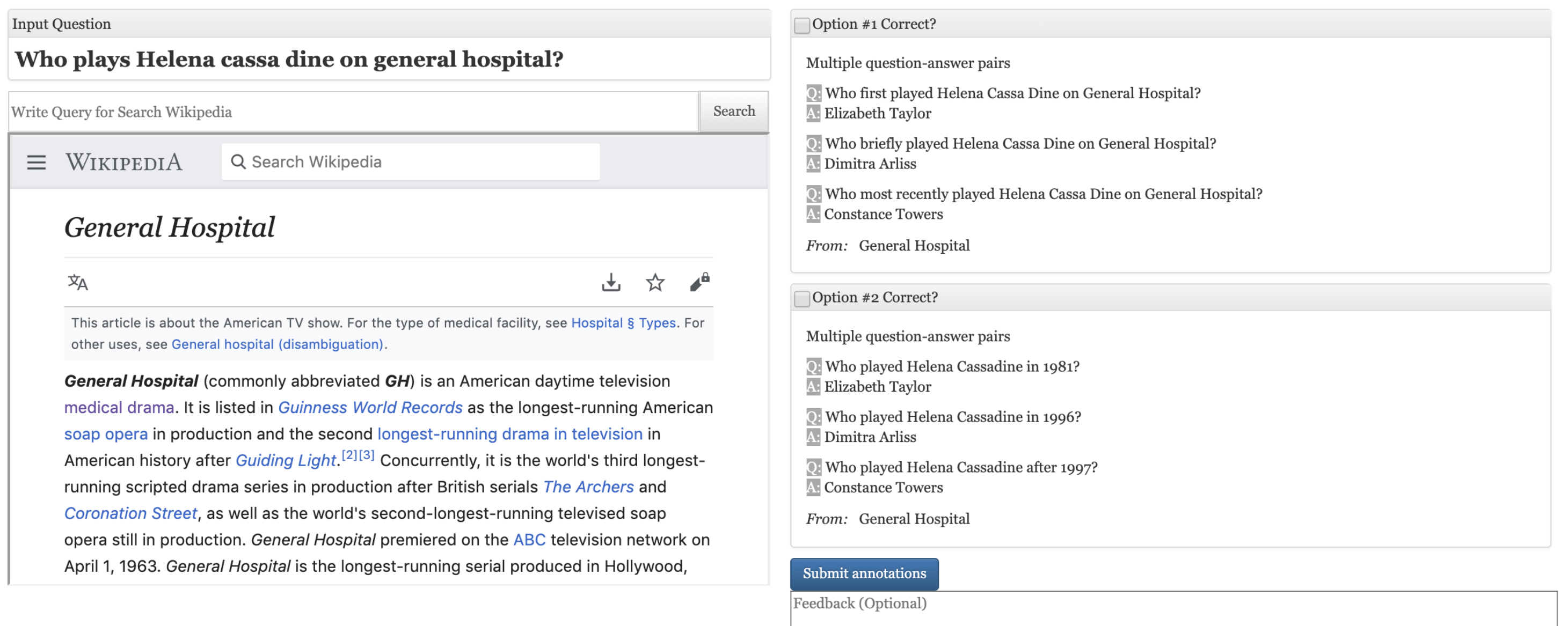}}\caption{
        Interface in the validation stage when the workers are given annotations from two generation workers and click the Wikipedia page that the generation workers have read.
    }\label{fig:interface-3}\end{subfigure}
\caption{
    Interface for crowdsourcing.
}
\label{fig:interface}\end{figure*}

\paragraph{Crowdsourcing details.} Figure~\ref{fig:interface} shows the interface used for generation and validation.
We use an iframe to render Wikipedia pages in a mobile view, in order to provide the document format that they are familiar with, rather than the plain text with no formatting.
When workers write the questions and the answers in the generation stage, we show appropriate error messages (e.g.
when the written question is the same as the \prompt) or warning messages (e.g., when the answer is composed of more than 20 words) in order to give tight feedback.
Workers produce free text answers which we instruct them to copy and paste from Wikipedia.

We pay 0.75 and 0.15 USD per \prompt\ for generation and validation, respectively.
Generators may skip the \prompt\ if the answer is not found in Wikipedia, or the question is ill-formed, too subjective or too ambiguous, e.g., ``When did the new tax cuts go into effect?"

\paragraph{Quality control.}
We only recruit full-time workers that are dedicated to our task.
We were able to recruit full-time workers by requiring the minimum number of HITs that can be achieved by working 40 hours a week.
We also host a public website for them to monitor the validated statuses, ask questions on examples that they do not understand the validated result, or claim on the validation which is incorrect in their opinion.
We found it very useful to communicate with workers, give feedback, and fix the incorrect annotations.

\paragraph{Inter-annotator agreement.}
When two independent generators are evaluated on the answer list from each other, they obtain 60.8 \Fanswer.
Specifically, for 76\% of questions, all annotations passed validation, either automatically because they exactly matched (37\%) or because they were both accepted by validators (39\%).
In the remaining 24\% of cases, one annotator missed a possible question-answer pair that the other one found, or included an invalid question-answer pair.

To assess validation quality, two co-authors annotated a random sample of 50 validations. The average \Fanswer\ between the co-authors and workers was 89.0\%.
\section{Discrepancy between \dev\ and test in \nqopen}\label{app:dev-and-test}In our experiments on \dataname, we found a significant discrepancy
between the \dev\ and test sets. Upon further investigation, we identified that this is at least in part due to a distributional difference between the \dev\ and test sets of \nqopen, upon which we built the data.
As this may be important for other researchers working on \nqopen, we detail our findings here.


Following \citet{lee2019latent}, \nqopen\ is constructed by filtering \nq\ to questions where at least one annotator provided a non-null short answer to the question.\footnote{
    \nq\ annotators answered each question with a set of short answers, which could be empty if there was no reasonable short answer. We refer to the empty cases as \textit{null answers}.
    See \citet{kwiatkowski2019natural} for details.
}
While the training and \dev\ sets of \nqopen\ were all drawn from the training set of \nq, in which one annotator answered each question, the test set of \nqopen\ is taken from its \dev\ set, which had five annotators per question.

This difference in number of annotators introduces a sampling bias: questions for which an annotator is less likely to find an answer are overrepresented in the \nqopen\ test set, in comparison to training and \dev.
Suppose, for example, that a randomly sampled annotator has a 50\% chance of producing a short answer for some question $q$. Then $q$ has a 50\% chance of making it into \nqopen's \devshort\ set, but a ($1 - .5^5 = $)~97\% chance of making it into test.
Concretely, when each annotator is considered independently, 34.6\% of the short answer annotations in the test set of \nqopen\ are null answers, and the majority of annotations are null for 33.9\% of questions. 


\begin{table}
    \centering \footnotesize
	\begin{tabular}{l P{0.6cm} P{0.6cm} P{0.6cm} P{0.6cm} }
    \toprule
        \multirow{2}{*}{Model} &
        \multicolumn{2}{c}{Any} & \multicolumn{2}{c}{First} \\
        \cmidrule(lr){2-3} \cmidrule(lr){4-5}
        & {dev} & {test} & {dev} & {test} \\
    \midrule
        \citet{min2019knowledge} & 34.7 & 34.5 & 32.4 & 25.7 \\ 
        \citet{asai2019learning} & 31.7 & 32.6 & 28.9 & 23.8 \\ 
        \citet{karpukhin2020dense} & 39.8 & 41.5 & 37.0 & 29.8 \\ 
        \modelname & 42.0 & 42.2 & 38.8 & 31.1 \\ 
    \bottomrule
    \end{tabular}
	\caption{
	    Exact Match (EM) on \nqopen\ of different models, counting a prediction as correct if it matches \textit{Any} gold reference, or only the \textit{First} non-null one.
	}\label{tab:dev-vs-test}
\end{table}

As a consequence, there is a significant gap in model performance between \devshort\ and test when they are evaluated under the same conditions.
The official evaluation protocol for \nqopen\ counts a prediction as correct if it matches any of the gold reference answers. Under these conditions, the gap between \devshort\ and test appears marginal (Table~\ref{tab:dev-vs-test}, first two columns).
However, as the \nqopen\ test set was more comprehensively annotated than \devshort, it has a more generous evaluation; the number of unique reference answers is 1.2 and 1.8 on \devshort\ and test, respectively.
In order to make the evaluation more consistent, we try evaluating models against the first reference answer only, and find a significant gap between \devshort\ and test (5--8\%) across all models (Table~\ref{tab:dev-vs-test}, last two columns).\footnote{
It is unlikely that this discrepancy is due to overfitting on \devshort, because the effect is consistent across models and not present on the other datasets that they are evaluated on.}

Despite this discrepancy, \dataname\ follows the setup and data split from \nqopen\, providing consistency with prior work.
Since the \dataname\ \dev\ and test sets were annotated under the same conditions, this discrepancy now shows up in the metrics.
We leave the distribution shift of questions on the test data as one of challenges on \dataname.

\section{Data Analysis Details}\label{app:data-analysis}

\paragraph{Mismatches with \nqopen.} 
29.4\% of \dataname\ \dev\ examples do not include the \nqopen\ answer.
We analyze a random sample of 50 such questions, and present a breakdown in Table~\ref{tab:comparison-to-nqopen-answer}.
We find that our answers are correct in 92\% of cases, among which 44\% of disagreements are due to mismatched spans, 22\% are due to the \nqopen\ answer being incorrect, and 14\% are due to time-dependence in the question. Of the 8\% of cases where our answer is incorrect, the \nqopen\ answers are also incorrect over half the time, indicating that these may be difficult questions.

\section{Baseline Implementation Details}\label{app:impl-details}\paragraph{Evidence corpus.} We use English Wikipedia dump from 2018-12-20 and 2020-01-20 for \nqopen\ and \dataname, respectively.
Following \citet{karpukhin2020dense}, we take the plain text and split passages to be up to 100 words each.

\paragraph{Model implementation.} All models are implemented in PyTorch~\citep{pytorch}, PyTorch-Transformers~\citep{Wolf2019HuggingFacesTS} (for \bert) and fairseq~\citep{ott2019fairseq} (for \bart). We use \textsc{BERT$_\text{base}$} and \textsc{BART$_\text{large}$} for all models.
We use the exact same setup and hyperparameters for any process that we follow \citet{karpukhin2020dense}.
For the passage retrieval through a dual encoder, we use the provided multi-setting trained model.
For all \bart-based models, we follow the default hyparameters from \bart\ summarization code in fairseq, using one 32GB gpu. For finetuning, we change the learning rate to be $5e-6$ on both tasks.
We use beam search for decoding the sequence.
We train the model for $4$ epochs (when trained on \nqopen\ or pseudo-labelled data) or $15$ epochs (when trained on \dataname), and take the best checkpoint based on the \dev\ data.
Note that the perplexity of the output sequence does not correlate with the metric of interest (Exact Match, \Fanswer\ or \Fedit) as briefly discussed in Section~\ref{subsec:main-results}, so using the metric of interest instead of perplexity is important for hyperparamter tuning or the choice of the best checkpoint. 

\paragraph{Details in ensemble and co-training.} We use an ensemble based on voting; the answers that are predicted by the highest number of models are chosen as the final answers.
The number of models used in ensemble ($C$) is $C=5$ before cotraining and $C=4$ after cotraining.
For co-training, we use $N=2$ and $C=6$, where $N$ is the number of iteration and $C$ is the number of models, in line with Algorithm~\ref{alg:cotraining}.
The choice of $C$ is determined by taking the best combination of the models as follows.
We train sixteen different models, using different hyperparamers including checkpoints from \nqopen, learning rates, the order of the answers in the output sequence and the random seed.
We then measure the \dev\ \Fanswer\ on different combinations of the models with varying $C$ ($4 \leq C \leq 6$) and take the best one.

\section{Error Analysis of \modelname}\label{app:err-analysis}
Table~\ref{tab:appendix-pred-analysis} reports an analysis of predictions by \modelname, on 50 random samples from the \dev\ set. We refer to Section~\ref{subsec:analysis} for the discussions.


\begin{table*}
    \centering
    \footnotesize
    \setlength\tabcolsep{3.8pt}
    \begin{tabular}{l}
    \toprule
        \textbf{Answer span mismatch (44\%)} \\
    \midrule
        {Q:} Who did the artwork for pink floyd's wall? \\
        {\nqopen\ answer:} Gerald Anthony Scarfe \\
        {\dataname\ answer:} \\
        Q: Who did the art work for the album cover of Pink Floyd's The Wall? / A: Gerald Scarfe \\
        Q: Who was the cinematographer for Pink Floyd - The Wall (1982 film)? / A: Peter Biziou \\
    \midrule
        \textbf{\nqopen\ answer incorporated as a question (2\%)} \\
    \midrule
        {Q:} What award did leonardo dicaprio won for the revenant? \\
        {\nqopen\ answer:} BAFTA Award; Academy Award for Best Actor; Golden Globe Award \\
        {\dataname\ answer:} \\
        Q: What British Academy Film Awards award did leonardo dicaprio won for the revenant? / A: Best Actor in a Leading Role \\
        Q: What Academy award did leonardo dicaprio won for the revenant? / A: Best Actor \\
        Q: What Golden Globe award did leonardo dicaprio won for the revenant? / A: Best Actor in a Motion Picture – Drama \\
        (Other question-answer pairs omitted) \\
    \midrule
        \textbf{\nqopen\ answer less specific (10\%)} \\
    \midrule
        {Q:}  When was the nba 3 point line introduced? \\
        {\nqopen\ answer:} 1979 \\
        {\dataname\ answer:} June 1979 \\
    \midrule
        \textbf{\nqopen\ answer incorrect and our answers include all possible answers (22\%)} \\
    \midrule
        {Q:} Who was inducted into the national inventors hall of fame first? \\
        {\nqopen\ answer:} John Fitch \\
        {\dataname\ answer:} Thomas Edison \\
        {\em Comment:} Thomas Edison inducted in 1973,  John Fitch inducted in 2006. John Fitch is mentioned as the earliest born  \\ inventor inducted.$^\dagger$
        \\
    \midrule
        \textbf{Mismatch from time-dependence (14\%)} \\
    \midrule
        {Q:} Who has the most home runs in the home run derby? \\
        {\nqopen\ answer:} Todd Frazier \\
        {\dataname\ answer:} \\
        Q: Who has the most home runs in the the TV show the home run derby? / A: Mickey Mantle; Mickey Charles Mantle \\ 
        Q: Who has the most home runs in the annual competition the home run derby? / A: Joc Russell Pederson; Joc Pederson \\
    \midrule
        \textbf{\nqopen\ answer is reasonable and our answers miss it (4\%)} \\
    \midrule
        {Q:} Who was the first person to settle dodge city? \\
        {\nqopen\ answer:} civilians \\
        {\dataname\ answer:} Henry J. Sitler \\
    \midrule
        \textbf{\nqopen\ answer incorrect but our answers miss another possible answer (4\%)} \\
    \midrule
        {Q:} In which year were chips used inside the computer for the first time? \\
        {\nqopen\ answer:} 1975 \\
        {\dataname\ answer:} 1962 \\
        {\em Comment:} The years that the chips were used for the first time in the prototype and the production are 1962 and 1974, \\ respectively, and can be both included.$^\ddagger$
         \\
    \bottomrule
    \end{tabular}
    \caption{
        Breakdown of cases that \nqopen\ answer is not included in \dataname\ answers. \\
        {\scriptsize $^\dagger$\url{en.wikipedia.org/wiki/List_of_National_Inventors_Hall_of_Fame_inductees}} \\
        {\scriptsize $^\ddagger$\url{en.wikipedia.org/wiki/History_of_computing_hardware_(1960s\%E2\%80\%93present)}}
    }
    \label{tab:comparison-to-nqopen-answer}
\end{table*}

\begin{table*}
    \centering
    \footnotesize
    \setlength\tabcolsep{3.8pt}
    \begin{tabular}{l}
    \toprule
        \textbf{Reference has multiple answers; Multiple answer prediction is correct (2\%)} \\
        \textbf{\Prompt:} Who was england's prime minister during ww1? \\
        \textbf{Reference:} H. H. Asquith ({\em beginning of WW1}), David Lloyd George ({\em end of WW1}) \\
        \textbf{Prediction:} (\Fanswer=1.00) H. H. Asquith, David Lloyd George \\
    \midrule
        \textbf{Reference has multiple answers; Multiple answer prediction is partially correct (40\%)} \\
        \textbf{\Prompt:} Who played kelly on the drew carey show? \\
        \textbf{\nqopen\ answer:} Cynthia Watros \\
        \textbf{Reference:} Cynthia Watros ({\em as Kellie N.}), Jenny McCarthy  ({\em as M. Kelly}), Brett Butler ({\em as G. Kelly}), Anna Gunn  ({\em as Kelly W.}) \\
        \textbf{Prediction:} (\Fanswer=0.40): Brett Butler \\
    \midrule
        \textbf{Reference has multiple answers; Multiple answer prediction is  incorrect (14\%)} \\
        \textbf{\Prompt:} Who plays the white queen in alice through the looking glass? \\
        \textbf{Reference:} Amelia Crouch ({\em young White Queen}), Anne Hathaway ({\em adult White Queen}) \\
        \textbf{Prediction:} (\Fanswer=0.00): Helena Bonham Carter$^\dagger$ \\
    \midrule
         \textbf{Reference has one answer; over-generated predictions (2\%)} \\
        \textbf{\Prompt:} How many times csk reached final in ipl? \\
        \textbf{Reference:} eight \\
        \textbf{Prediction:} (\Fanswer=66.7): eight, seven$^\ddagger$ \\
    \midrule
        \textbf{Reference has one answer; correct single answer prediction (26\%)} \\
        \textbf{\Prompt:} When did the 5th circuit became the 11th circuit? \\
        \textbf{Reference:} October 1, 1981 \\
        \textbf{Prediction:} (\Fanswer=100.0): October 1, 1981 \\
    \midrule
        \textbf{Reference has one answer; incorrect single answer prediction (12\%)} \\
        \textbf{\Prompt:} Who is considered the home team for super bowl 52?  \\
        \textbf{Reference:} New England Patriots \\
        \textbf{Prediction:} (\Fanswer=0.0): Atlanta Falcons \\
    \midrule
        \textbf{Reference is incorrect (4\%)} \\
        \textbf{\Prompt:} Who has won the most trophies man utd or liverpool? \\
        \textbf{Reference:} Man utd ({\em trophies}), Liverpool ({\em FIFA and UEFA Cups}) \\
        \textbf{Prediction:} (\Fanswer=66.7): Manchester United \\
    \bottomrule
    \end{tabular}
    \caption{
        Analysis of multiple answer predictions made by \modelname\ with co-training, on 50 samples from the \dev\ data.
        Rewrites are omitted but differentiation of multiple answers is denoted as a keyword in {\em italic}. \\
        {\footnotesize $^\dagger$Helena Bonham Carter played Red Queen.} \\
        {\footnotesize $^\ddagger$In fact, the model may have found time-dependency, because the eighth event happened only in 2019.} 
    }
    \label{tab:appendix-pred-analysis}
\end{table*}

\end{document}